\documentclass{article}

\usepackage[final, nonatbib]{nips_2018_ml4health}
\usepackage[numbers]{natbib}

\usepackage[utf8]{inputenc} 
\usepackage[T1]{fontenc}    
\usepackage{hyperref}       
\usepackage{url}            
\usepackage{booktabs}       
\usepackage{amsfonts}       
\usepackage{nicefrac}       
\usepackage{microtype}      
\usepackage{amsmath}
\usepackage{graphicx}  
\usepackage{multirow}
\usepackage{subfig}
\usepackage{tikz}
\usetikzlibrary{bayesnet}
\usetikzlibrary{arrows,decorations.markings}
\usepackage[title]{appendix}

\DeclareMathOperator*{\KL}{KL}
\DeclareMathOperator*{\ent}{\mathcal{H}}
\DeclareMathOperator{\ELBO}{\mathcal{L}}

\title{Semi-unsupervised Learning of Human Activity using Deep Generative Models}

\author{
  Matthew Willetts $^{1,2}$, Aiden Doherty$^{1}$, Stephen Roberts$^{1,2}$, Chris Holmes$^{1,2}$ \\
$^{1}$ University of Oxford\\
$^{2}$ The Alan Turing Institute
}

\begin{document}
\setlength{\abovedisplayskip}{3pt}
\setlength{\belowdisplayskip}{3pt}
\maketitle

\begin{abstract}
We introduce `semi-unsupervised learning', a problem regime related to transfer learning and zero-shot learning where, in the training data, some classes are sparsely labelled and others entirely unlabelled. Models able to learn from training data of this type are potentially of great use as many real-world datasets are like this. Here we demonstrate a new deep generative model for classification in this regime. Our model, a Gaussian mixture deep generative model, demonstrates superior semi-unsupervised classification performance on MNIST to model M2 from Kingma and Welling (2014). We apply the model to human accelerometer data, performing activity classification and structure discovery on windows of time series data.

\end{abstract}

\section{Introduction}
\label{intro}

While developing machine learning solutions in healthcare and medicine, the amount of unlabelled data is typically much larger than the amount of labelled data. Further, there is selection bias: the labelled data is often from a biased sample of the overall data distribution. For rare diseases, rare risk factors and other uncommon states, particular class categories might be entirely unobserved in the labelled dataset, only appearing in unlabelled data. 

This occurs in the use of machine learning to measure sleep duration and physical activity from sensor data to better understand their association with morbidity and mortality \cite{Doherty2017, Willetts2018}. By increasing our understanding of human activity, combined with rich medical datasets such as biobanks, we can potentially obtain deeper insights into diseases and their links with lifestyle factors.

Thus we are interested in the case where an unlabelled instance of data could be from one of the sparsely-labelled classes or from an entirely-unlabelled class. We call this `semi-unsupervised learning'. Here we are jointly performing semi-supervised learning on sparsely-labelled classes, and unsupervised learning on completely unlabelled classes. We give a deep generative models \cite{Rezende2014, Kingma2013} that can solve this problem.\footnote{Note: related work-in-progress, with more focus on theory, is in the NeurIPS Bayesian Deep Learning Workshop 2018, titled `Semi-unsupervised Learning using Deep Generative Models'.}

Semi-unsupervised learning has similarities to some varieties of zero-shot learning (ZSL), where deep generative models have been of interest \cite{Weiss2016}, but in ZSL one has access to auxiliary side information (commonly an `attribute vector') for data at training time, which we do not. So our regime is equivalent to transductive generalised ZSL, but with no side information \cite{Xian2018}. It also has similarities to transfer learning. In Cook et al.'s terms \cite{Cook2013}, `semi-unsupervised learning' is related to uninformed semi-supervised transductive transfer learning but here we use our source and target information jointly, and our discrete label space can either be the same or different for our labelled and unlabelled data.

Numerous methods have been proposed for activity recognition, including: deep discriminative models \cite{Alsheikh2015, Ordonez2016, Hammerla2016}; random forests \cite{Ellis2016}; probabilistic graphical models/Hidden Markov Models \cite{Ellis2016, Willetts2018}; and Gaussian processes \cite{Alvarez2011}. None are set up with the `semi-unsupervised' regime in mind. Thus we are interested in flexible, scalable semi-unsupervised machine learning methods, where unlabelled and labelled data are used together to improve classification performance. We show our model's utility for MNIST image data classification, and for human activity recognition from sensor data.

\section{Deep Generative Models}

In a deep generative model, the parameters of the distributions within a probabilistic graphical model are themselves parameterised by neural networks. The simplest is a variational autoencoder \cite{Kingma2013, Rezende2014}, the deep version of factor analysis. Here there is a continuous unobserved latent $z$ and observed data $x$. The joint probability is $p_{\theta}(x,z)=p_{\theta}(x|z)p(z)$ with $p(z)=\mathcal{N}(0,\mathbb{I})$ and $p_{\theta}(x|z) = \mathcal{N}(\mu_{\theta}(z),\Sigma_{\theta}(z))$ where $\mu_{\theta}(z),\Sigma_{\theta}(z)$ are each parameterised by neural networks with parameters $\theta$. As exact inference for $p(z|x)$ is intractable, it is standard to perform stochastic amortised variational inference to obtain an approximation $q(z|x)$ to the true posterior.

For a VAE, introduce a recognition network $q_\phi(z|x) = \mathcal{N}(\mu_{\phi}(x),\Sigma_{\phi}(x))$ (where $\mu_{\phi}(z),\Sigma_{\phi}(z)$ are neural networks with parameters $\theta$). Through joint optimisation over $\{\theta, \phi\}$ using stochastic gradient descent we aim to find the point-estimates of the parameters $\{\theta,\phi\}$ that maximises the evidence lower bound $\ELBO(x)= - \KL_{z}(q_\phi(z|x) || p_{\theta}(x,z))$. For the expectation over $z\sim q_{\phi}(z|x,y)$ in $\ELBO(x)$ we take Monte Carlo (MC) samples. To then take derivatives through these samples wrt $\theta, \phi$ we need to be able to differentiate through a sample from a Gaussian. For this we use a case of the `reparameterisation trick' where we notice that is possible to rewrite a sample from a Gaussian as a deterministic function of sample from $\mathcal{N}(0,\mathbb{I})$:
\begin{equation}
z\sim\mathcal{N}(\mu,\sigma^2) \Longleftrightarrow \epsilon\sim\mathcal{N}(0,\mathbb{I}), z=\mu + \sigma \cdot \epsilon
\end{equation}
thus we can differentiate a sample w.r.t. $\mu, \sigma^2$, so we can differentiate our MC approximation w.r.t $\theta, \phi$.

To perform semi-supervised classification with a deep generative model, introduce a discrete class variable $y$ into the generative model and into the recognition networks. There will be two evidence lower bounds for the model, one where $y$ is a latent variable to be inferred: $\ELBO(x)=- \KL_{z,y}(q_\phi(z, y|x) || p_{\theta}(x,y,z))$ and one where $y$ is observed: $\ELBO(x, y)=- \KL_{z}(q_\phi(z|x, y) || p_{\theta}(x,y,z))$.\\

In this work we build on the M2 model developed by Kingma and Welling (2014) \cite{Kingma2014a}. Here $p_{\theta}(x,y,z)= p_{\theta}(x|y,z)p(y)p(z)$ and $q_{\phi}(z,y|x)=q_{\phi}(z|y,x)q_{\phi}(y|x)$, $q_{\phi}(y|x)=\textrm{Cat}(\pi_{\phi}(x))$ and $q_{\phi}(z|x,y)=\mathcal{N}(\mu_{\phi}(x,y),\Sigma_{\phi}(x,y))$. $p(y)$ is the discrete prior on $y$. Via simple manipulation one can show $\ELBO(x)= \sum_y [q_{\phi}(y|x) \ELBO_\ell(x,y)] +\ent(q_{\phi}(y|x))$. Note that $q_{\phi}(y|x)$, which is to be our trained classifier at the end, only appears in $\ELBO(x)$, so it would only be trained on unlabelled data. To remedy this, motivated by considering a Dirichlet hyperprior on $p(y)$, they add to the loss the cross entropy between the true label and $q_{\phi}(y|x)$, weighted by a factor $\alpha$. So the overall loss for model M2 with unlabelled data $D_u$ and labelled data $D_\ell$ is:
\begin{equation}
\ELBO(D_u, D_\ell) = \sum_{x_u \sim D_u} \ELBO(x_u) +\sum_{(x_\ell,y_\ell) \sim D_\ell} \Big[\ELBO(x_\ell,y_\ell) + \alpha  (- y_\ell \cdot \log q_{\phi}(y|x))\Big]
\end{equation}
This model has been demonstrated in the semi-supervised case \cite{Kingma2014a}, but when there is no label data at all, when we are just optimising $\sum_{x^u \sim D^u} \ELBO(x^u)$, the model can fail to learn an informative distribution for $q_{\phi}(y|x)$ (see similar effect in \cite{Dilokthanakula}). Either it collapses to the prior $p(y)$, or it maps every datapoint to one class. Either way the model reduces to something very similar to a standard VAE with no $y$ variable. This happens when the encoder and decoder are high enough in capacity to obtain a locally optimal value of the evidence lower bound without using the class label. Thus, if one wishes to use high-capacity neural networks it is necessary to adjust the model in some way. Here we propose a change to the generative structure.
\begin{figure}[h!]
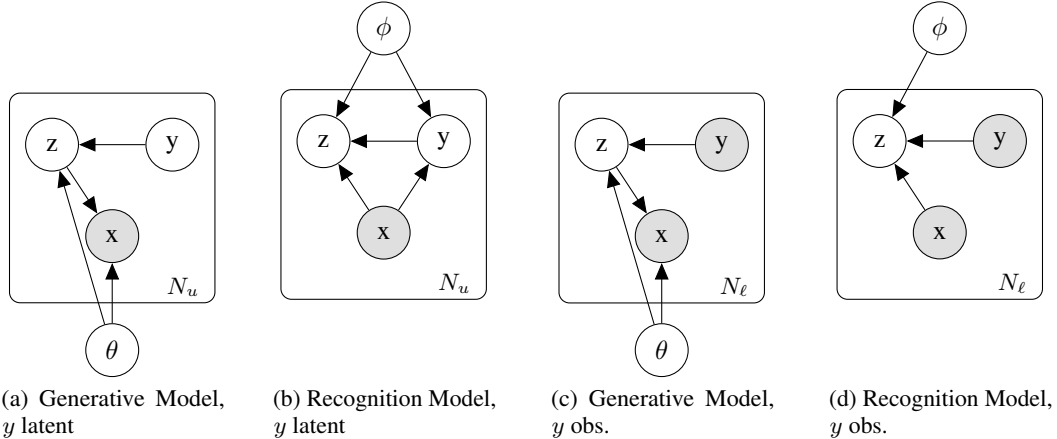

\centering
\noindent\makebox[\textwidth]{%
\subfloat[][Generative Model, $y$ latent]{
\tikz{ %
      \node[obs] (x) {x} ; %
    \node[latent, above=of x, xshift=-0.8cm, yshift=-0.5cm] (z) {z} ; %
    \node[latent, above=of x, xshift=0.8cm, yshift=-0.5cm] (y) {y} ; %
    \node[latent, below=of x, , yshift=0.2cm] (theta) {$\theta$} ; %
    \plate[inner sep=0.2cm, xshift=-0cm, yshift=0.12cm] {plate1} {(z) (x) (y)} {$N_u$}; %
    \edge {theta} {z, x} ; %
    \edge {z} {x} ; %
        \edge {y} {z} ; %

  }
  \label{fig:VAEgenerative1}}
  \hspace*{6mm}
\subfloat[][Recognition Model, $y$ latent]{
\raisebox{6.5ex}
{\tikz{ %
      \node[obs] (x) {x} ; %
    \node[latent, above=of x, xshift=-0.8cm, yshift=-0.5cm] (z) {z} ; %
    \node[latent, above=of x, xshift=0.8cm, yshift=-0.5cm] (y) {y} ; %
    \node[latent, above=of x, yshift = 1cm] (phi) {$\phi$} ; %
    \plate[inner sep=0.2cm, xshift=-0cm, yshift=0.12cm] {plate1} {(z) (x) (y)} {$N_u$}; %
    \edge {phi} {z, y} ; %
    \edge {y} {z} ; %
        \edge {x} {y,z} ; %

  }}
  \label{fig:VAErecognition2}}
    \label{fig:VAEgenerative3}
  \hspace*{6mm}
\subfloat[][Generative Model, $y$ obs.]{
\tikz{ %
      \node[obs] (x) {x} ; %
    \node[latent, above=of x, xshift=-0.8cm, yshift=-0.5cm] (z) {z} ; %
    \node[obs, above=of x, xshift=0.8cm, yshift=-0.5cm] (y) {y} ; %
    \node[latent, below=of x, yshift=0.2cm] (theta) {$\theta$} ; %
    \plate[inner sep=0.2cm, xshift=-0cm, yshift=0.12cm] {plate1} {(z) (x) (y)} {$N_\ell$}; %
    \edge {theta} {z, x} ; %
    \edge {z} {x} ; %
        \edge {y} {z} ; %

  }
  \label{fig:VAErecognition4}}
    \label{fig:VAEgenerative5}
  \hspace*{6mm}
\subfloat[][Recognition Model, $y$ obs.]{
\raisebox{6.5ex}
{\tikz{ %
      \node[obs] (x) {x} ; %
    \node[latent, above=of x, xshift=-0.8cm, yshift=-0.5cm] (z) {z} ; %
    \node[obs, above=of x, xshift=0.8cm, yshift=-0.5cm] (y) {y} ; %
    \node[latent, above=of x, yshift = 1cm] (phi) {$\phi$} ; %
    \plate[inner sep=0.2cm, xshift=-0cm, yshift=0.12cm] {plate1} {(z) (x) (y)} {$N_\ell$}; %
    \edge {phi} {z} ; %
    \edge {y} {z} ; %
        \edge {x} {z} ; %
  }}
  \label{fig:VAErecognition6}}
}  
  \caption{Representation of our DGM as a probabilistic graphical model, for data $x$, partially observed class $y$, continuous latent $z$, $\theta$, $\phi$. Figures (a,c) shows the generative model $p_\theta(x|z)p_\theta(z|y)p(y)$ with $y$ latent and observed. Figures (b,d) shows the variational approximate posterior $q_\phi(z,y|x)$ with $y$ latent and observed.}
    \label{fig:VAE}
\end{figure}

M2's inability to consistently learn $y$ in the semi-unsupervised case is not satisfactory for us as we are interested in cases when some classes of data are not found in our labelled dataset. This is to enable us to learn with both semi-supervised and unsupervised classes. Many deep generative models have been proposed for semi-supervised learning \cite{Maaloe2016, Maaloe2017a} and for unsupervised learning \cite{Dilokthanakula, Burda, Kingma2016}, but none have dealt with posterior collapse in $y$ so as to perform semi-unsupervised learning. So we propose a deep generative model: a Gaussian mixture version of a variational auto-encoder, inspired by Kingma et al.'s M2 \cite{Kingma2014a} and the GMM-VAE \cite{Dilokthanakula}. Rather than having the same distribution $p(z)$ for all classes as in M2, we condition on $y$ to obtain a mixture of gaussians in $z$ space. We perform semi-unsupervised classification with this model, and also compare performance with M2. Note that our model can also be trained unsupervised. The generative model for the data is:
\begin{align}
p_{\theta}(x,y,z)&=p_{\theta}(x|z)p_{\theta}(z|y)p(y)\\
p(y)&= \textrm{Cat}(\pi)\\
p_{\theta}(z|y)&= \mathcal{N}(\mu_{\theta}(y),\sigma^2_{\theta}(y))\\
p_{\theta}(x|z)= \mathcal{N}(&\mu_{\theta}(z),\sigma^2_{\theta}(z)) \textrm{ or } \mathcal{B}(\mu_{\theta}(z))
\end{align}
We then perform amortised stochastic variational inference, with variational distributions as before for M2. See Figure 1 for a graphical representation of our model.

\section{Experimental setup}

\subsection{Model Implementation}
All networks are small MLPs, 2-4 layers with 500 hidden units per layer and RELU activations. $z$ is 100 dimensional. The same network architectures were used for networks with the same inputs and outputs. Our code is based on the template code associated with Gordon $\&$ Hernandez-Lobato (2017) \cite{Gordon2017}. Training was done using Adam \cite{Kingma2015}. Kernel initialisation was Glorot-Normal and weights were regularised via a Gaussian prior as in \cite{Kingma2014a}. \footnote{Code accompanying the paper is available at: \href{https://github.com/MatthewWilletts/GM-DGM}{github.com/MatthewWilletts/GM-DGM}.} 

\subsection{MNIST experiment}
Here we trained the both our model and M2 with digits ${0,1,2,8,9}$ semi-supervised with $100$ labels, and digits ${3,4,5,6,7}$ entirely unsupervised. We augmented $y$ with 5 extra classes to learn into in addition to the 5 vacated classes. $p(y)$ was equal to 1/10 for the 5 semi-supervised classes and 1/20 for each of the 10 unsupervised classes. To calculate accuracy we attributed the learnt, unsupervised classes to the most common class within it at test time. See Figure 2 for the resulting confusion matrices.

\subsection{Activity recognition experiment}
We trained and evaluated the model to distinguish between activity states using the CAPTURE-24 dataset. This dataset was gained from 132 people (mean age=42) wearing a wrist-mounted accelerometer for 2-3 days each. This device measured acceleration at 100Hz tri-axially along axes pinned to the device, saturating at $\pm8g$ \cite{Doherty2017}. We preprocess the data into 126 standard features extracted in 30 second windows as in Willetts et al. (2018) \cite{Willetts2018}. In addition to wearing an accelerometer, labels were obtained via wearable cameras that took photos every $\approx 20 sec$ \cite{Doherty2013, Gurrin2014}. These images were then classified visually by medical students into the classes of the Compendium of Physical Activity (\cite{Ainsworth2011}), which are detailed but overlap. Thus we select only the most unambiguous CPA classes (eg. \textit{17250: Walking as a means of transport} or \textit{1010: Bicycling}) to use as labels, the rest are considered unlabelled. See appendix for details. Further we only will give ourselves access to labels for a subset of 20 participants, the remaining 92 treated as entirely unlabelled with a test set of 20 participants. We treat the problem as classification of these windows -- but only having ground truth labels for a fraction of data. This dataset is also highly unbalanced. The rarest class, running, is $\sim 80$ times less prevalent than the commonest, sleep.
\section{Results}

\begin{figure}[h!]
\centering
\noindent\makebox[\textwidth]{%
\subfloat[][M2]{
\includegraphics[width = 5cm]{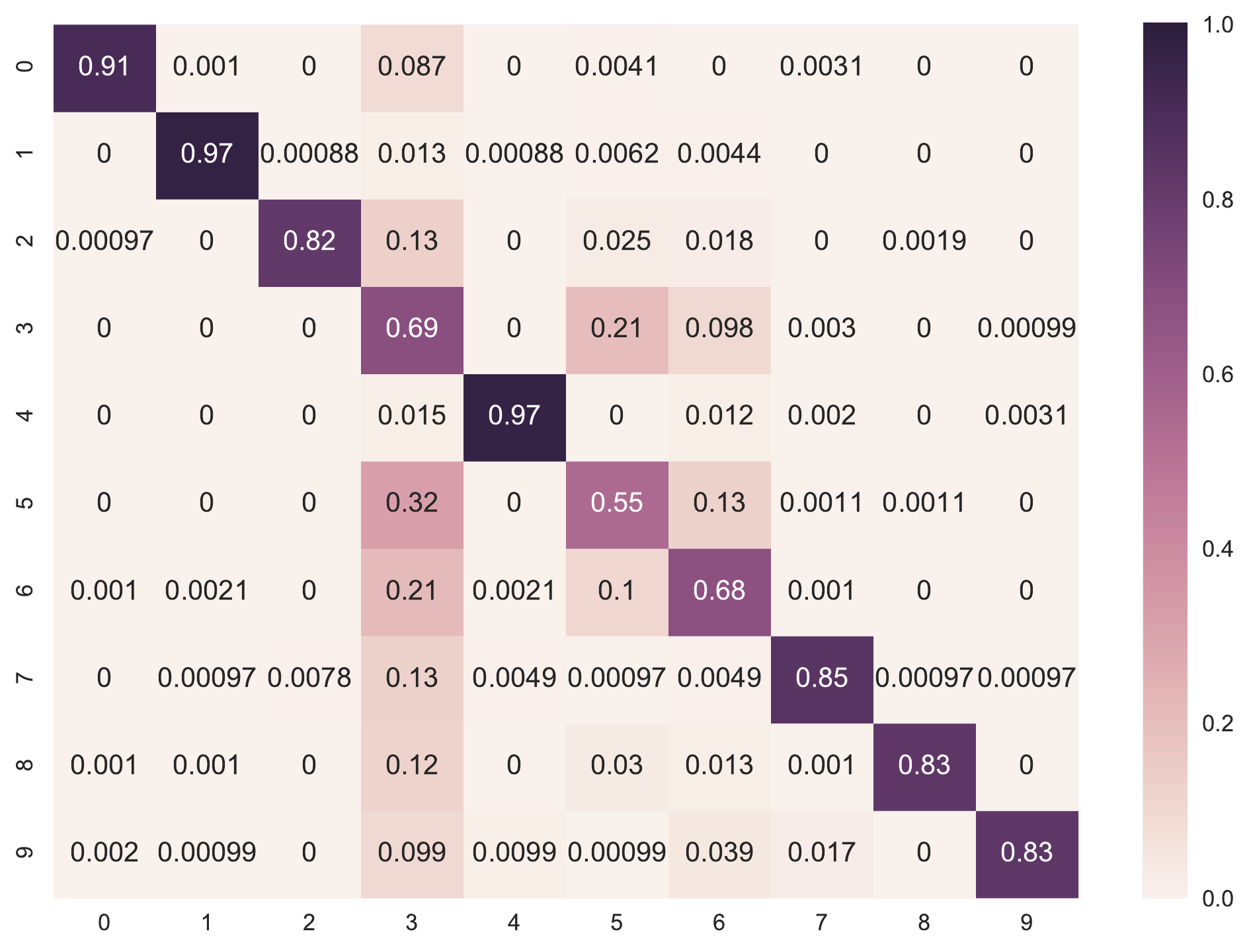}
  \label{fig:MNIST_m2}}
  \hspace*{0mm}
\subfloat[][Our Model]{
\includegraphics[width = 5cm]{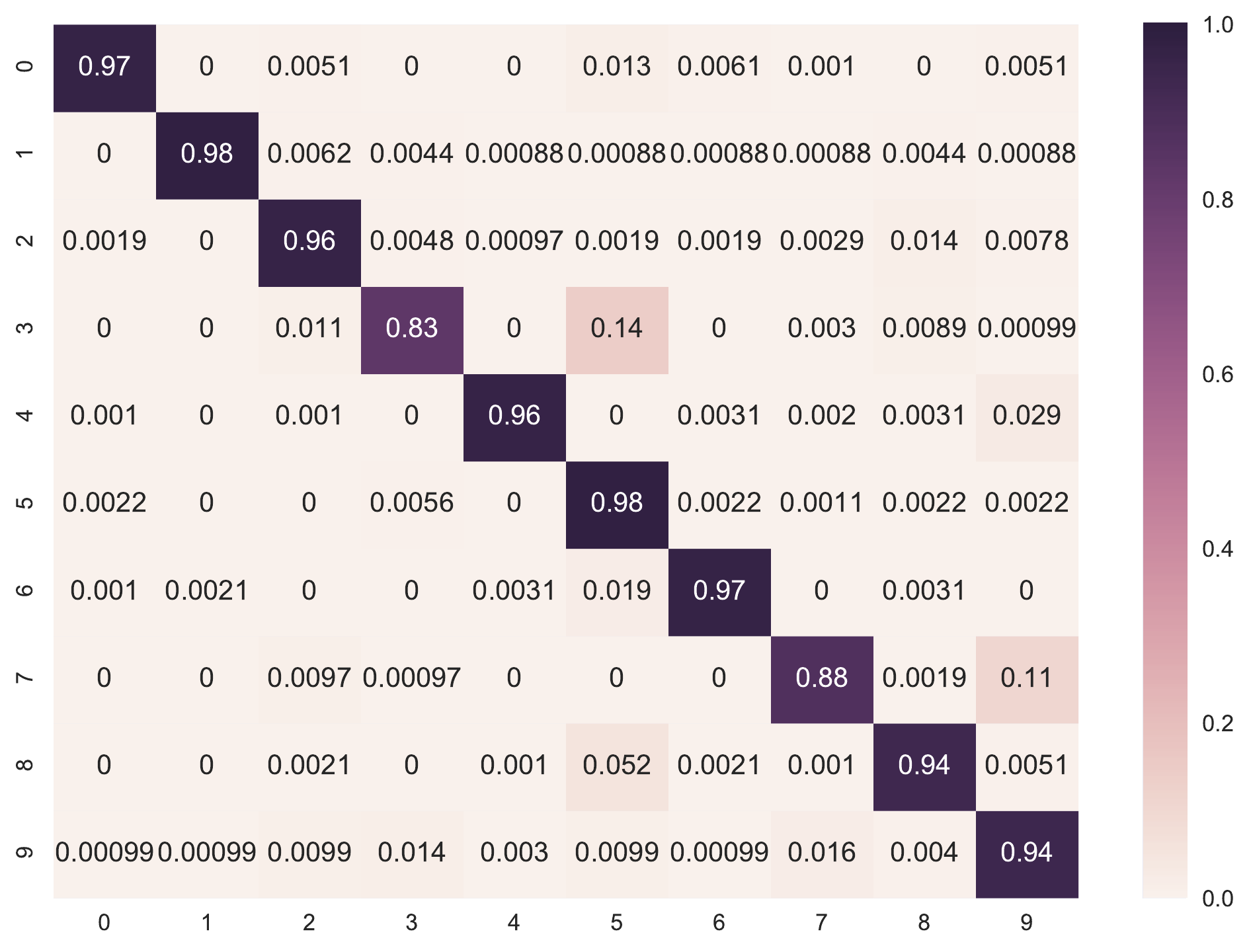}

\label{fig:MNIST_gm_dgm}}
}
\caption{Confusion matrices for a) M2 - accuracy 0.82 - and b) Our model - 0.94 - when trained on MNIST with 100 labelled examples each for digits 0,1,2,8,9 and digits 3,4,5,6,7 entirely unsupervised. 5 additional classes were added to $y$. All unsupervised classes were attributed to the true 10 classes by taking them to represent their most prevalent original class. Clearly the original model M2 shows more confusion between unsupervised classes than our model. Each are the best of 10 runs.}
\end{figure}
\vspace{-0.5cm}
\subsection{Activity Results}
Model performance is not well captured by simple scores on the predictions, as we are interested in state discovery so some data may be counted as an error for having been place in a new class when truly it should be. Looking at the the full CPA annotations within the discovered clusters, we see the model learns a class for activities while standing, containing: \textit{5035: kitchen activity general cooking, 11795: walking on job and carrying light objects, 11600: (generic) standing tasks such as store clerk, 11475: (generic) manual labour'} and a cluster for sitting activities: \textit{11580: office work/computer work general, 11580: office work such as writing and typing, 7030: sleeping, 9030:  sitting desk work}. Compared to M2, the overall Calinski-Harambasz score \cite{Calinski1974} rose from $\approx160$ to $\approx700$.

\section{Conclusion}
We show that our method can perform better than Kingma et al's M2 \cite{Kingma2014a} in semi-unsupervised learning on MNIST, and can discover structure in CAPTURE-24 data. $y$ and $z$ can be thought of as separating out class and style information about data $x$. Our model, through having a mixture of Gaussians in $z$ space is a suitable choice of model when different classes in data might have different stylistic information for different classes.
This is work in progress, the next steps are to apply these methods, with more flexible and powerful parameterisations of the parameters of the distributions, to the full UK Biobank Activity Dataset of 100k people using CAPTURE-24 as the small labelled dataset, and extend this to time series analysis.

\vspace{0.75in}

\newpage

\begin{appendices}
\section{Dictionary from CPA to Classes}
\normalsize{We chose a small number of well-represented and unambiguous CPS codes to form the sparsely-labelled classes in the activity data analysis. In total these CPS codes cover $\approx 34\%$ of the labelled data.}

\begin{table}[h!]
 \centering
  \caption{CPA Dictionary}
  \label{cpa-table}
  \centering
  \begin{tabular}{llll}
    \toprule
    Class     & CPA Code     & CPA Description & Proportion ($\%$)\\
    \midrule
    Bicycling & 1010  & Bicycling &1.5    \\
    Driving     & 6206 & Driving automobile or light truck & 3.4   \\
    Eating     & 13030 & eating sitting alone or with someone & 4.2  \\
    Riding Transport & 16016 & Riding in a bus or train & 1.5 \\
    Running & 12150 & Running & 0.12 \\
    Sitting & 9060 & Sitting/lying reading or without observable/identifiable activities & 8.2\\
    Sleep & 7030 & Sleep & 9.6 \\
    Walking & 17250 & Walking as the single means to a destination not to work or class & 2.5\\ 
    Walking & 17161 & Walking not as the single means of transport & 1.8\\
    Walking & 17270 & Walking as the single means to work or class& 0.55 \\
    Walking & 17082 & Hiking or walking at a normal pace through fields and hillsides& 0.18\\
    \bottomrule
  \end{tabular}
\end{table}

\end{appendices}

\end{document}